\def\year{2019}\relax
\documentclass[letterpaper]{article} 
\usepackage{aaai19}  
\usepackage{times}  
\usepackage{helvet}  
\usepackage{courier}  
\usepackage{url}  
\usepackage{graphicx}  
\usepackage[table]{xcolor}    

\usepackage[utf8]{inputenc} 
\usepackage[T1]{fontenc}    
\usepackage{url}            
\usepackage{booktabs}       
\usepackage{amsfonts}       
\usepackage{nicefrac}       
\usepackage{microtype}      

\usepackage{algorithm}

\usepackage{algorithmic}

\begin{document}
%
\title{Learning Latent Beliefs and Performing Epistemic Reasoning for Efficient and Meaningful Dialog Management}
\author{Aishwarya Chhabra, Pratik Saini, Amit Sangroya, C. Anantaram\\
TCS Innovation Labs, Tata Consultancy Services Limited, ASF Insignia, Gwal Pahari, Gurgaon, India\\
\texttt{(aishwarya.chhabra, pratik.saini, amit.sangroya, c.anantaram)@tcs.com}
}
\maketitle
\begin{abstract}

Many dialogue management frameworks allow the system designer to directly define belief rules to implement an efficient dialog policy. Because these rules are directly defined, the components are said to be hand-crafted. As dialogues become more complex, the number of states, transitions, and policy decisions becomes very large. To facilitate the dialog policy design process, we propose an approach to automatically learn belief rules using a supervised machine learning approach. We validate our ideas in Student-Advisor conversation domain, where we extract latent beliefs like student is \textit{curious, confused and neutral}, etc. Further, we also perform epistemic reasoning that helps to tailor the dialog according to student's emotional state and hence improve the overall effectiveness of the dialog system. Our latent belief identification approach shows an accuracy of 87\% and this results in efficient and meaningful dialog management.
 
\end{abstract}

\section{Introduction}
Dialog based interaction between a customer and a bot may become tedious and irrelevant if the system’s beliefs are not consistent with the set of beliefs of the customer. Beliefs are cognitive representational states that represent the presumed context of the conversation perceived by each agent. A key limitation in today's conversational interfaces is their lack of robustness when understanding the latent beliefs. The inherent difficulties of conversational systems (chatbots) are further increased by the conditions under which these systems typically operate: increasingly larger vocabularies, large and diverse user populations, spontaneous input, etc. Unless mediated by better belief identification and robust reasoning mechanisms, these errors propagate to subsequent stages of processing in a dialog system and exert a considerable negative impact on the quality and ultimately the success of the interaction.

Identifying latent beliefs is a challenging task since it depends on various factors like identifying and analyzing context. Other challenges that exist in identifying latent beliefs is the reference to multiple concepts and the need to extract a large amount of facts, commonsense knowledge, anaphora resolution, and logical reasoning. Consider the example, \textit{``I am looking for a heavy course"}, it is difficult to identify the latent belief if the student is looking for a course which is heavy in terms of the workload; class size and/or the nature/quality of assignments.

We propose a novel approach for identifying more accurate beliefs over concept values in conversational systems by integrating information across multiple turns in the conversation. Traditional machine learning approaches train a system with extremely large dialog corpus that covers a variety of scenarios. Another approach is to build a system with a complex set of hand-crafted rules that may address some specific instances. Both approaches may be impractical in many real-world domains. Our framework is based on a combination of machine-learning mechanism and logical reasoning to understand the latent beliefs. Our model then evaluates the beliefs to tailor the dialog and make it consistent with the set of beliefs of the user. This process then helps drive the conversation in a meaningful way.

\section{Related Work}

Deep learning based dialog systems~\cite{miller2016key} use memory networks to learn the underlying dialog structure and carry out goal-oriented dialog. 
However, they do not factor in beliefs or trigger epistemic rules in modifying the conversation given the evolving context. 
In~\cite{williams2016dialog} Williams et.al, describe the dialog state tracking challenge and mention ``how the task of correctly inferring the state of the conversation - such as the user's goal - given all of the dialog history up to that turn'' is important. It is in this overall context, we propose that it is important to evaluate the probable beliefs held by the human and tailor the dialog system suitably to be consistent with the beliefs in order to hold a relevant conversation.

Although a number of attempts have been made to build dialog systems~\cite{henderson2014word}, ~\cite{Weston16}, ~\cite{williams2016dialog}, the use of epistemic rules in driving the dialog in a consistent way with the beliefs has not yet been tackled. Various approaches to dialog management have been proposed and these can be broadly classified into finite-state methods, probabilistic methods and deep learned methods. 

There are recent motivating examples of works that make use of machine-learning to build intelligent dialog systems.
Traditional dialog systems are specialized for a domain and rely on slot-filling driven by a knowledge base and a finite-state model~\cite{lemon2006hierarchical}. The finite-state model represents the dialog structure in the form of a state transition network in which the nodes represent the system's interactions and the transitions between the nodes determine all the possible paths through the network.

Uckelman~\cite{sara2010obligationes} describes how in a formal dialog system, dynamic epistemic logic can be used in an \textit{Obligatio}, where two agents, an \textit{Opponent} and a \textit{Respondent}, engage in an alternating-move dialog to establish the consistency of a proposition. 
Sadek et al.~\cite{Sadek:1997:AND:1622270.1622305} proposes a reasoning engine to build effective and generic communicating  agents.
Motivated by the above works, we propose to identify the beliefs, use these beliefs to trigger epistemic rules, and use the assertions of the rules to drive the conversation by tailoring the states in a finite-state machine dialog system.

Opinion mining methods have been in use for over long time.  Recently, these methods have been applied to dialog systems. Roy et al.~\cite{aaai_Roy_16} propose a novel approach to consider customer satisfaction to tailor the dialog. While general sentiment analysis methods are useful to understand a customer's mental situation, they may need to be complemented with more domain specific information leading to richer fine grained classes of sentiments.
For this reason, it is necessary to develop methods which take into account domain specific sentiments information.

\section{Supervised Approach for Automatically Learning Latent Beliefs}

A key issue in dialogue management is the design of the dialogue policy incorporating latent beliefs. While it is possible to design this by hand, such an approach has several shortcomings. Development is manually intensive and expensive, the systems are difficult to maintain and performance is often sub-optimal. In our previous work, we highlighted some of the issues in hand crafting belief rules~\cite{DBLP:conf/aaai/SangroyaASR18}.

\begin{figure}
\centering
      \includegraphics[scale=0.2]{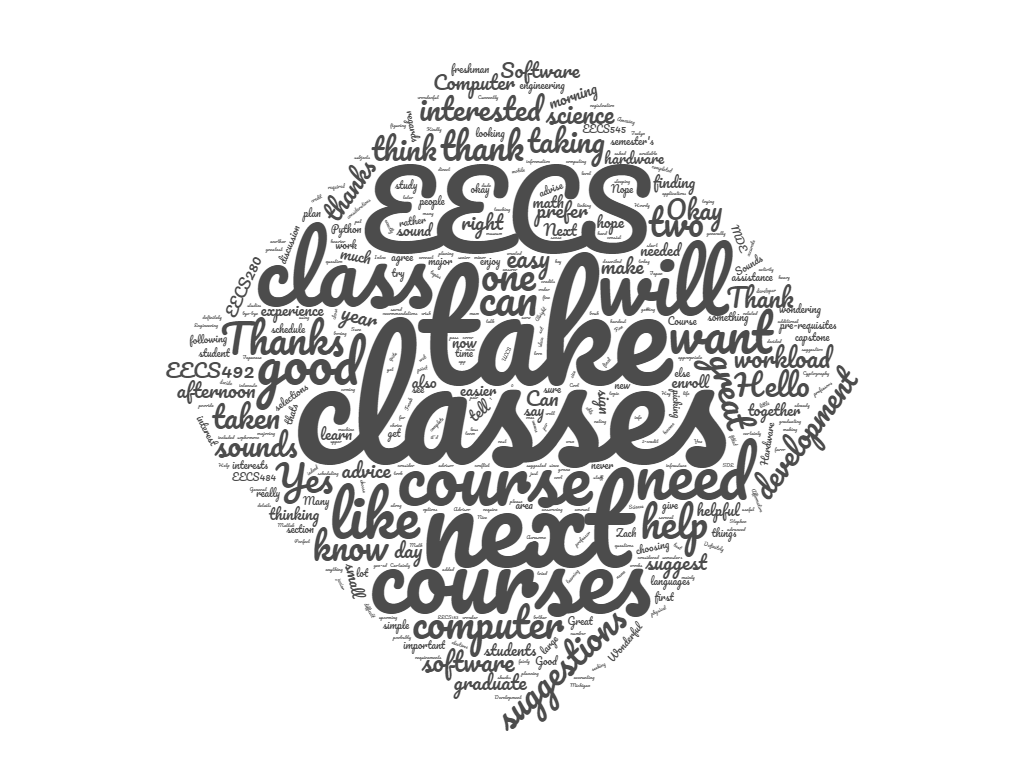}
    \caption{Word Cloud of Dialog Utterances of a \textit{Curious} Student}
    \label{word_cloud}
\end{figure}

An alternative to hand-crafting belief rules is to automatically learn them given a large annotated corpus of utterances and corresponding labeled beliefs. However, it is challenging to design a machine learning model as the performance of the model would be degraded if it is based on simple linguistic features such as \textit{bag-of-words} (See Figure~\ref{word_cloud}). Sometimes, the available training data consist of large number of closely related classes that makes the problem of belief identification more difficult. 

It is challenging task to build an automatic system that can understand the latent beliefs of humans (more specifically students). For example, there may be a student who is confused and directionless, might come up with something like \textit{``I have no inkling of where I want my life to go and am unable to determine what classes to take. I'm interested in machine learning. Can you help me decide what to do?”} . There may be another student who has done his homework on what he wants but needs only little direction from the advisor. These two different categories of students needs different conversational flow and a different set of dialog policy needs to be tailored for each case. 

We take the initial input from the student to advisory bot and categorize it into one of the tree categories based on there emotional belief. We consider 3 categories ``\textbf{curious}",  ``\textbf{neutral}", ``\textbf{confused}" and manually annotate them to student's utterances for each dialog. Most of the conversations seem to be similar in nature. Therefore, it becomes a challenging task to annotate the data correctly and pass it to machine learning model. 

\begin{figure}
\centering
      \includegraphics[scale=0.25]{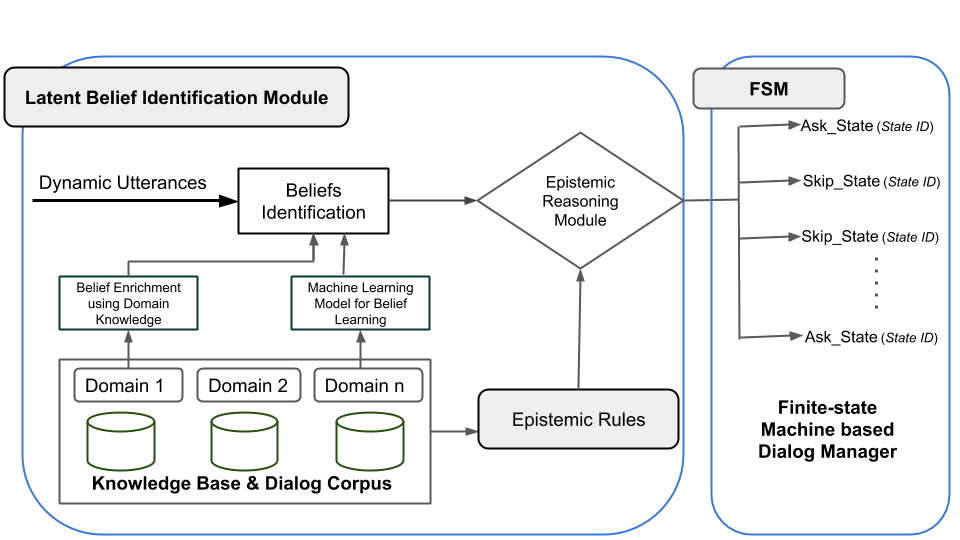}
    \caption{Domain Specific Latent Belief Extraction Process}
    \label{fig:senti_methodology}
\end{figure}

Figure~\ref{fig:senti_methodology} illustrates the proposed methodology for identifying and updating the latent beliefs. Its inputs include the utterances and a domain ontology.

\subsection{Step 1: Classifying Latent Beliefs}

If an utterance belongs to more critical categories such as \textit{confused}, it is assigned a higher weight as compared to category such as \textit{curious}. This is intuitive that student with a more confused state of mind would need more attention and a specific dialog plan. The latent belief classification is done with the help of machine learning in an automatic fashion.

\subsection{Step 2: Enriching Beliefs using Domain Ontology}

\begin{figure}
\centering
      \includegraphics[scale=0.3]{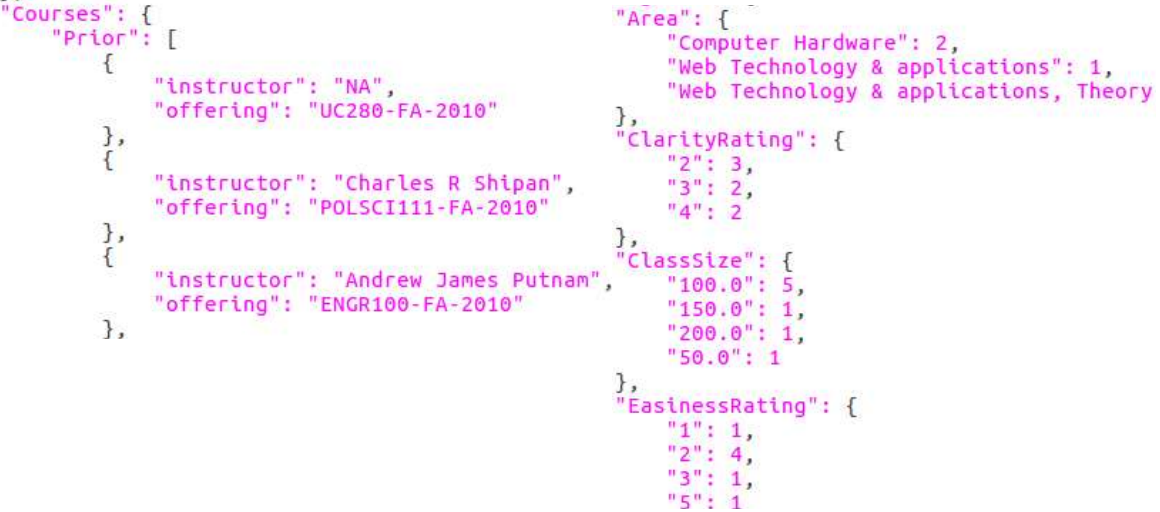}
    \caption{Domain Knowledge for Belief Enrichment}
    \label{fig:meta_info}
\end{figure}

We make use of knowledge mining derived using the domain ontology to update the latent beliefs. Our ontology consists of a large knowledge graph expressing the information about a domain. For example, this includes the course information, their easiness ratings, workload ratings and class size (See Figure~\ref{fig:meta_info}). This information is available in a structured RDF graph. Use of ontology assists in enriching the beliefs further. It also helps to map the utterance which can be in any natural language form to a structured information that can be used to tailor the FSM based dialog manager. 

\subsection{Epistemic Reasoning over Latent Beliefs and Domain Knowledge}

The extracted latent beliefs and the domain knowledge trigger the epistemic rules. For example ``\texttt{Belief (Student is confused) and Course\_Load(High) => Knows-Agent (student is not confident), Knows-Agent (should advise light courses)}'' asserts facts about the current epistemic state of the agent. We use \texttt{Prolog} to define epistemic rules and carry out the reasoning against facts and domain knowledge.

\subsection{Step 3: Learning the Model for Latent Belief Extraction}

We now train a LSTM (Long Short Term Memory) network to build a model that can automatically categorize the beliefs based upon the information described in previous section (See Algorithm~\ref{alg1}). Like many other studies of LSTM on text, words are first converted to low-dimensional dense word vectors via a word embedding layer. The first layer is therefore the embedding layer that uses 32 length vectors to represent each word. The next layer is the LSTM layer with 100 units. Finally, we use a dense output layer with 5 neurons (5 classes/labels) and a softmax activation function to make the predictions. We used \textit{Categorical\_Cross\_Entropy} as the loss function (in \textit{Keras}) alongwith ADAM optimizer. For regularization, we employ dropout to prevent co-adaptation. We run the experiment for 20 epochs with a batch size of 64.

\begin{algorithm} 
\caption{Algorithm for Updating Dialog States using Latent Belief Extraction} 
\label{alg1} 
\begin{algorithmic} 

   \REQUIRE FSM for dialog   
   \REQUIRE User Input as Dialog Utterance $T$
   \ENSURE Fine Grain Set of Latent Beliefs
   \ENSURE Meaningful dialog states
   \FOR{all Utterances $r$ in $T$ }
   \FOR{all states ${s_i}$ of FSM}
   \STATE Remove Stopwords and Punctuations
   \STATE Convert to Lowercase 
   \STATE Tokenize
   \STATE Mark special name-entities
   
   \FOR{all sentence $m$ in $r$ }
   \STATE Extract Latent Beliefs {$Curious$, $Confused$ etc.}
   \STATE Using Domain Ontology {$courses$, $workload$, $size$ etc., update latent beliefs $b$}
   \STATE Using Information Extraction, update $b$
   \STATE Using facts ${F_i}$ and EPISTEMIC\_rulebase, fill slots of Dialog manager Frames
   \STATE update weight ${w_i}$ of state ${s_i}$ using Latent Beleifs and Epistemic Reasoning
   \STATE Using threshold and weights, build meaningful states of FSM (ask\_states, skip\_states)
    \ENDFOR
    \ENDFOR
    \ENDFOR

\end{algorithmic}
\end{algorithm}

\begin{figure}
\centering
     \includegraphics[scale=0.45]{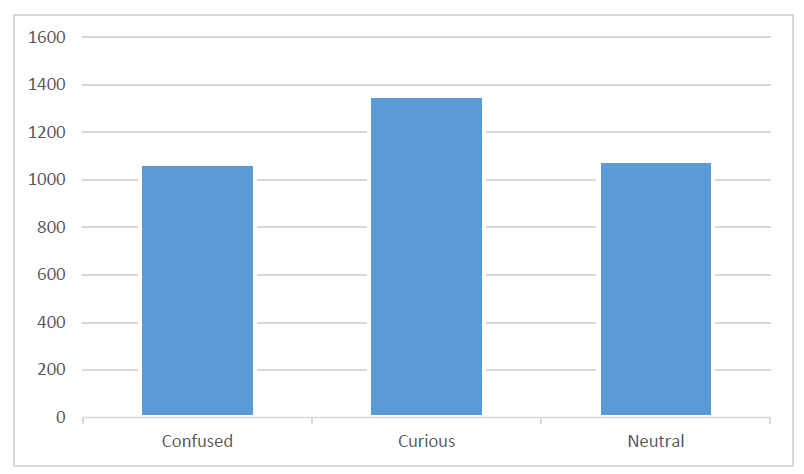}
    \caption{Number of Utterances in Training Data for each Category (Latent Beliefs)}
    \label{fig:belief_classes}
\end{figure}

In our experiments we consider dialog dataset student advisor conversations from \url{https://www.ibm.com/blogs/research/2018/07/sentence-selection-dstc7}. We manually annotate latent beliefs across three categories such as \emph{Curious, Confused and Neutral}. Total number of utterances after data processing were 3500 (Figure~\ref{fig:belief_classes}). The clean-up process involves converting text to lower case; tokenizing the sentences; and removal of punctuations and stopwords. Each input utterance as input is converted into a vector form. We identify the top 300 unique words and every word in this vocabulary is given a index. If the word is not present in vocabulary we consider it as 0. For Example: \textit{I am very disappointed today}. The vector representation of this sentence is [10, 100, 23, 467, 0]. Next, we need to truncate and pad the input sequences so that they are all the same length. We take the max length of utterance to be 50.  We divide the data into training set (75\%)  and test set (25\%). We ensure that we do not have data sparsity issue i.e. We keep approximate equal proportion of data for each class.

\subsection{Experimental Results}

We evaluate the proposed method on the seventh Dialog State Tracking Challenge (DSTC-7) dataset (``Flex Data: Student – Advisor dialogue"~\cite{Flex}). We used 3500 conversations including list of paraphrases for each utterance as training data for belief classification. It becomes a little challenging to tag them to correct categories because of lexical similarity between these three classes. We got 84\% classification accuracy for latent belief extraction. We also tried bag-of-words and naive bayes models that performed badly for this experiment. As we discussed earlier, this is possible due to lot of common words between three classes.

%

\subsection{Example: Tailoring the Chatbot}
\label{chatbot}

We parse a dialog through Dependency parsers (such as Stanford-CoreNLP, GATE, MITIE etc.) and
extract triples from the description by focusing on the dependencies identified among nouns and verbs. For example, for a description \textit{''I prefer morning classes as I sleep early at night"}, triples such as (\textit{I, prefer, morning classes}) are extracted. Once the triples are extracted, we use a hand-crafted fact-assertion rulebase to assert facts implied by the triples. This is done by evaluating the triples in the context of a student-advisor ontology, synonym-and-slang dictionary, information-extraction patterns that are relevant for the category of the belief, and by triggering the fact-assertion rules. The latent belief identification process helps to understand the student's emotional beliefs and tailor the conversations accordingly.

We assume that we have a hand-crafted dialog-management finite-state-machine (FSM) to carry out the dialog with the student. The FSM operates on slots that are filled by extracting information from the input dialog and subsequent interaction. The probable beliefs of the student that were asserted as facts are then evaluated by the epistemic rules encoded in a knowledge base for the domain. The rules make assertions about the states in the FSM that need to be skipped and the states that need to be evaluated in order to be consistent with the beliefs of the student. The subsequent dialog is carried out and the next set of beliefs are then asserted. The cycle then continues. As shown in Table~\ref{Sample_Output}, we can observe that chatbot is able to skip some FSM states as a result of latent beliefs. The beliefs and epistemic rules helped tailor the dialog to the user's expectations. In this work, we demonstrate the overall architecture where we use the output of RNN based belief identification model as an input to the epistemic rule engine. Our approach is generic and can be applied easily in any other domains.

  \rowcolors{2}{gray!25}{white}

\begin{table}
\footnotesize
\begin{center}
\caption{Sample Output of Dialog System}
\label{Sample_Output}
\begin{tabular}{|p{3.2cm}|p{4.1cm}|}
    \rowcolor{gray!50}

\hline
Advisor & Hi! I am your advisor. You can ask any doubts in selection of your courses for next semester.
\\ 
\hline 

\hline 
Student & I am a junior year student with interest in statistics and data analysis
\\
\textcolor{blue}{Output ($ML$-$Model$)} & \texttt{\textcolor{blue}{Latent Belief: Student(Curious)}}\\

\hline
Advisor \textcolor{red}{(skipstate(ask\_interest), skipstate(ask\_semester))} & Do you have any specific requirement \textit{about the workload} of the course
\\

\hline
Student & I would prefer a \textit{class with lighter workload} and higher helpfulness rating
\\

\hline
Advisor & Do you have any timing preferences?
\\

\hline
Student & I \textit{prefer morning classes} as I sleep early at night.\\

\hline
Advisor \textcolor{red}{(SkipState(ask\_extra\_details))} & I would advise you STATS250 ``Statistics and Data Analysis" which is an easy course.\\

\hline

\end{tabular}
\end{center}
\end{table}

\section{Conclusion}
This paper presents a supervised approach to identify latent beliefs from dialog utterances and tailor the dialog. In future, we would like to investigate reinforcement learning based approaches for belief identification.

\bibliographystyle{aaai} \bibliography{CHI_2017}

\end{document}